\pdfoutput=1

\documentclass[11pt]{article}

\usepackage{EACL2023}
\usepackage{graphicx}
\usepackage{times}
\usepackage{latexsym}
\usepackage{booktabs}
\usepackage{amsmath}
\usepackage[T1]{fontenc}

\usepackage[utf8]{inputenc}

\usepackage{microtype}

\usepackage{inconsolata}

\title{MIReAD: simple method for learning high-quality representations from scientific documents}

\author{Anastasia Razdaibiedina$^{\diamondsuit,\spadesuit}$ \and Alexander Brechalov$^\diamondsuit$ \\ 
$^\diamondsuit$University of Toronto \and $^\spadesuit$Vector Institute \\
\texttt{anastasia.razdaibiedina@mail.utoronto.ca} \\\texttt{alexander.brechalov@utoronto.ca} 
}

\begin{document}
\maketitle
\begin{abstract}
Learning semantically meaningful representations from scientific documents can facilitate academic literature search and improve performance of recommendation systems. Pre-trained language models have been shown to learn rich textual representations, yet they cannot provide powerful document-level representations for scientific articles. We propose \textsc{MIReAD}, a simple method that learns high-quality representations of scientific papers by fine-tuning transformer model to predict the target journal class based on the abstract. We train \textsc{MIReAD} on more than 500,000 PubMed and arXiv abstracts across over 2,000 journal classes. We show that \textsc{MIReAD} produces representations that can be used for similar papers retrieval, topic categorization and literature search. Our proposed approach outperforms six existing models for representation learning on scientific documents across four evaluation standards. \footnote{MIReAD model weights are available through HuggingFace at \href{https://huggingface.co/arazd/MIReAD}{https://huggingface.co/arazd/MIReAD} 
\\
\\
Abstracts and journal data is available through HuggingFace Hub at \href{https://huggingface.co/datasets/brainchalov/pubmed_arxiv_abstracts_data}{https://huggingface.co/datasets/brainchalov/\\pubmed\_arxiv\_abstracts\_data}
\\
\\
Full code is provided at \href{https://github.com/arazd/miread}{https://github.com/arazd/miread}} 
\end{abstract}

\section{Introduction}

A significant increase in the volume of scientific publications over the past decades has made the academic literature search a more challenging task. One of the key steps to improve the recommendation systems (RS) for research articles is to obtain high-quality document-level representations. Recently, transformer-based models have brought substantial progress to the field of natural language processing (NLP), obtaining state-of-the-art results on a variety of benchmarks \cite{vaswani2017attention, devlin2018bert}. While transformer models are effective in language modeling and learning sentence representations, deriving document-level representations for scientific articles remains a challenge.

\begin{figure}
\centering
\includegraphics[scale=0.3]{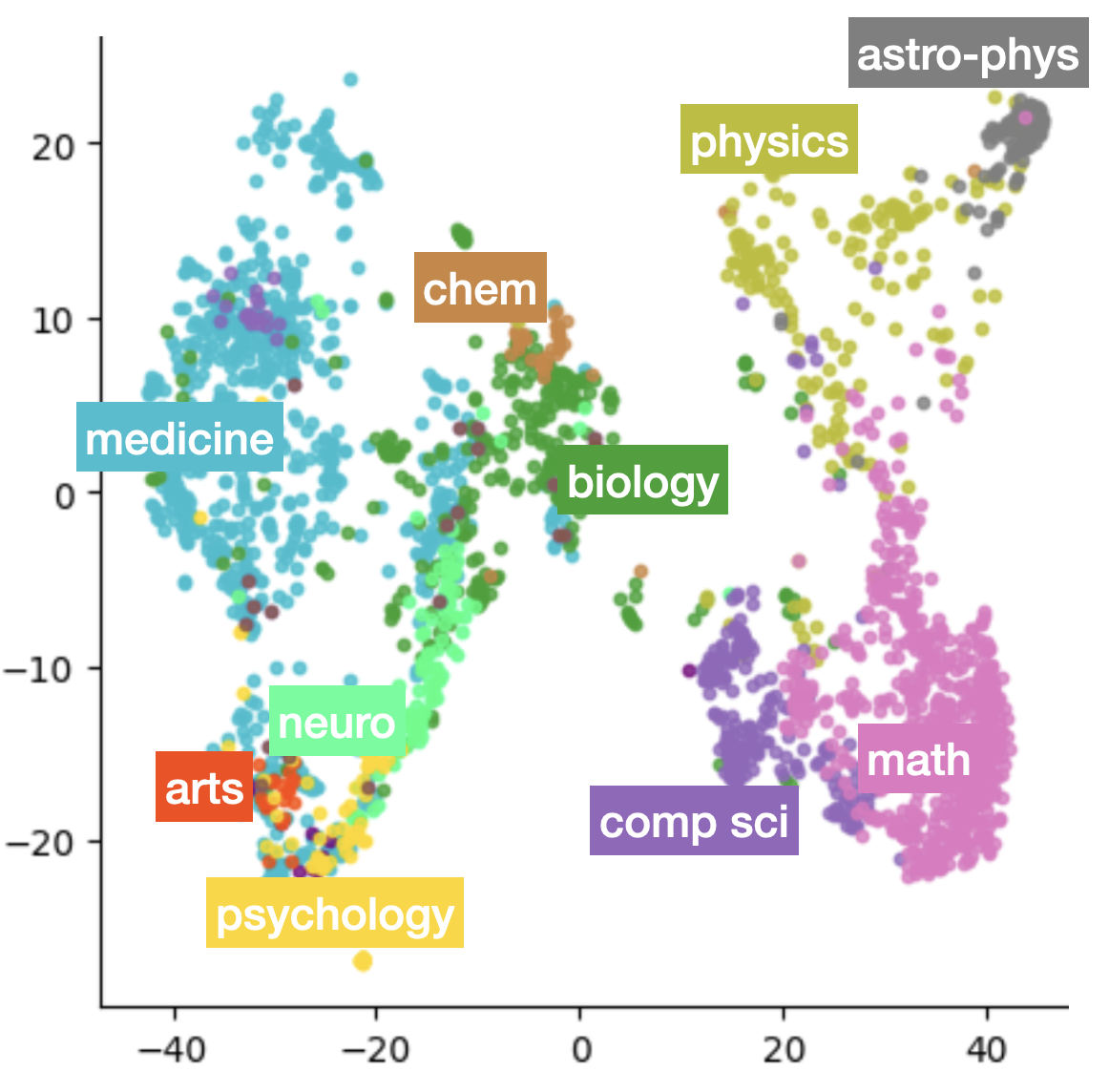}
\caption{\textsc{MIReAD} representations allow distinguishing abstracts by scientific domain, without using domain or citation information during finetuning. tSNE with abstracts' representations from unseen journals is shown.
 }
\label{fig:method}
\end{figure}

Previous transformer-based methods for representation learning on scientific documents are derived from BERT model \cite{devlin2018bert}. Classic examples of such approaches are PubMedBERT, BioBERT and SciBERT - scientific domain adaptations of BERT, which were pre-trained with masked language modeling (MLM) objective on PubMed abstracts, as well as full-text articles from PubMedCentral and Semantic Scholar, respectively \cite{pubmedbert, lee2020biobert, beltagy2019scibert}. While MLM objective allows to efficiently capture the context of the sentence, it cannot achieve accurate paper representations that can be used "off-the-shelf" to discover similar articles.
To address this problem, recent works explored fine-tuning the pre-trained models with supervised objectives based on citation graphs \cite{wright2021citeworth, cohan2020specter}. 
Despite their efficiency, citation-based objectives have several disadvantages: (1) citations are not distributed uniformly, with novel papers and articles from certain fields being less favoured; (2) citations have a bias related to the increased self-citation and existence of over-cited papers; (3) citation graphs are often large and difficult to preprocess. Hence, there is a gap in representation learning for scientific articles, requiring approaches which would derive high-quality document-level representations, without relying on the citation graphs. 

In this paper, we propose \textbf{\textsc{MIReAD}}, an approach that requires \textbf{M}inimal \textbf{I}nformation for \textbf{Re}presentation Learning of \textbf{A}cademic \textbf{D}ocuments. \textsc{MIReAD} combines the SciBERT architecture with novel training objective - a target journal classification. We show that such a simple training objective leads to high-quality representations of academic papers, suitable for RS usage. Figure~\ref{fig:method} illustrates how \textsc{MIReAD} representations from unseen abstracts are separated based on scientific domain, even though this information was not accessed during training. We trained \textsc{MIReAD} by predicting one of 2,734 journal classes from the paper's title and abstract for 500,335 articles from PubMed and arXiv. Then we measured the quality of paper representations obtained with \textsc{MIReAD} using three evaluation standards - linear evaluation, information retrieval, and clustering purity scores - on three different datasets. \textsc{MIReAD} substantially outperforms 5 previous approaches (BERT, PubMedBERT, BioBERT, SciBERT, CiteBERT) across all evaluation benchmarks and outperforms SPECTER in most cases.

\section{Methods}

\subsection{\textsc{MIReAD}}
\textsc{MIReAD} is based on BERT architecture and we initialize it from SciBERT's weights. 
We fine-tune \textsc{MIReAD} to predict journal class solely from paper’s abstract and title with cross-entropy loss:
\[ L(\widehat{y}_i, y_i) = - \sum_{i=1}^{N} y_i \log(\widehat{y}_i)  \]
Here $\widehat{y}_i$ and $y_i$ stand for predicted probability and ground truth label of the class $i$, $N$ is equal to 2734, the total number of unique journal classes.


\textsc{MIReAD} takes as input a concatenation of paper's title and abstract, appended to the {\fontfamily{qcr}\selectfont
[CLS]} token, and separated by the {\fontfamily{qcr}\selectfont
[SEP]} token: 
\[\text{input} = \text{{\fontfamily{qcr}\selectfont
[CLS]}} \text{title} \text{{\fontfamily{qcr}\selectfont
[SEP]}} \text{abstract} \]
Final paper representation $v$ is obtained by passing the input through the transformer model, and taking the representation of the {\fontfamily{qcr}\selectfont
[CLS]} token:
\[v = \text{forward}(\text{input})_{\text{{\fontfamily{qcr}\selectfont
[CLS]}}}\]

\subsection{Dataset}
To achieve a good coverage of different knowledge domains, we constructed a dataset from arXiv and PubMed abstracts and their corresponding metadata (title and journal) \cite{clement2019arxiv}. We limited the number of abstracts per each journal to not exceed 300, and excluded journals with less than 100 abstracts or no publications in year 2021. The final dataset contains 500,335 abstracts (181,967 from arXiv and 318,368 from PubMed), covers 76 scientific fields and 2,734 journals. More details on dataset preparation are in Appendix~\ref{label:data_preparation}. We fine-tune \textsc{MIReAD} for one epoch on all paper abstracts using 1e-6 learning rate.


\subsection{Baseline models}
We compare \textsc{MIReAD} to six baseline approaches based on BERT \cite{devlin2018bert}. We use the original BERT model, its three different domain adaptations: BioBERT \cite{lee2020biobert}, PubMedBERT \cite{pubmedbert} and SciBERT \cite{beltagy2019scibert}, as well as two representation extraction models trained with citation objectives: CiteBERT \cite{wright2021citeworth} and SPECTER \cite{cohan2020specter}. Additionally, we include SentenceBERT \cite{reimers-2019-sentence-bert} -- a modification of the BERT model that includes siamese network structure to find semantically similar sentence pairs.


\begin{table*}[t]
\centering
\fontsize{10}{11}\selectfont
\setlength{\tabcolsep}{4pt}
\begin{tabular}{lllllllll}
\toprule
Task $\rightarrow$ & \multicolumn{2}{c}{MAG} & \multicolumn{2}{c}{MeSH} & \multicolumn{2}{c}{arXiv \& PubMed} & \multicolumn{2}{c}{Unseen journals} \\
    Model $\downarrow$ & \multicolumn{1}{c}{F1} & \multicolumn{1}{c}{Acc.} & \multicolumn{1}{c}{F1} & \multicolumn{1}{c}{Acc.} & \multicolumn{1}{c}{F1} & \multicolumn{1}{c}{Acc.} & \multicolumn{1}{c}{F1} & \multicolumn{1}{c}{Acc.} \\
\midrule
      BERT & $71.47_{0.82}$ & $77.82_{0.49}$ & $46.33_{0.41}$ & $63.67_{0.33}$ &  $4.22_{0.21}$ & $22.05_{0.92}$ &  $2.62_{0.19}$ &   $4.70_{0.25}$ \\
PubMedBERT & $72.65_{0.97}$ &  $78.25_{0.4}$ &  $72.45_{0.8}$ &  $77.80_{0.51}$ &  $4.07_{0.27}$ &  $19.00_{0.73}$ &  $0.71_{0.31}$ &   $1.41_{0.5}$ \\
   BioBERT & $59.43_{0.22}$ & $71.63_{0.38}$ &  $50.60_{1.12}$ & $67.87_{0.58}$ &   $2.53_{0.2}$ &  $20.00_{0.84}$ &  $0.65_{0.22}$ &  $1.95_{0.44}$ \\
   SciBERT & $74.84_{0.57}$ & $79.47_{0.35}$ & $66.67_{0.98}$ & $74.19_{0.58}$ & $10.75_{0.71}$ &  $31.30_{0.45}$ &   $7.90_{1.38}$ & $11.46_{1.64}$ \\
  CiteBERT & $70.49_{0.58}$ &  $76.40_{0.21}$ & $55.94_{1.23}$ &   $67.80_{0.30}$ &  $9.26_{0.49}$ & $29.05_{1.07}$ &  $6.72_{0.73}$ & $10.19_{0.79}$ \\
  SentBERT$^*$ &  $80.5$ &  $-$ & $69.1$ &   $-$ &  $-$ & $-$ &  $-$ & $-$ \\
   SPECTER & $81.47_{0.18}$ & $\mathbf{85.05_{0.14}}$ & $86.23_{0.27}$ & $87.38_{0.13}$ & $30.75_{0.69}$ & $44.92_{0.49}$ & $18.26_{1.34}$ & $23.73_{1.17}$ \\
    \textsc{MIReAD} & $\mathbf{81.85_{0.59}}$ & $84.85_{0.31}$ & $\mathbf{86.71_{0.36}}$ & $\mathbf{88.22_{0.19}}$ &  $\mathbf{34.97_{0.3}}$ & $\mathbf{48.95_{0.26}}$ & $\mathbf{19.35_{0.49}}$ & $\mathbf{25.11_{0.36}}$ \\
\bottomrule
\end{tabular}
\caption{Linear evaluation of document-level representations obtained from different methods. We report F1-score and accuracy on four standards. Mean and standard deviation across three runs is shown. $^*$ denotes results reported by \cite{cohan2020specter}.}
\label{tab:1}
\end{table*}

\section{Evaluation of representations}
We evaluate the information content of the representations of scientific abstracts produced by different approaches. Ideally, we are interested in representations that contain information about scientific domain, and allow to distinguish specific subdomains within larger fields. We use three common strategies for representation quality assessment: linear evaluation, clustering purity and information retrieval.  

\subsection{Linear evaluation of representations}
We first evaluate representations with commonly used \textit{linear evaluation protocol} \cite{zhang2016colorful, oord2018representation, chen2020simple}. Under this protocol, a linear classifier is trained on top of extracted representations, and test accuracy is used as a quality metric. Hence, better information content of the representations translates into higher classification accuracy. Details of training the logistic regression are provided in Appendix~\ref{label:linear_probing}.
In our experiments, we perform linear evaluation of representations derived from the abstracts from four datasets, with varying degree of difficulty: 

\textbf{Academic topics}
In this task, we predict the research field of the paper using Microsoft Academic Graph (MAG) dataset \cite{sinha2015overview}. MAG provides paper labels, which are organized into a hierarchy of 5 levels. We follow SciDocs evaluation framework by \citet{cohan2020specter}, which provides a classification dataset with labels from level 1 topics (e.g. business, sociology, medicine etc.), and has a train-test split. Overall, MAG dataset consists of 19 classes and covers 25K papers.

\textbf{Medical subject headings}
We use Medical Subject Headings (MeSH) dataset by  \citet{lipscomb2000medical} to classifiy academic paper representations into one of 11 disease classes (e.g. diabetes, cardiovascular disease etc.). Similarly to MAG, we use data with train and test splits provided by SciDocs. This dataset contains a total of 23K medical papers.

\textbf{PubMed and arXiv categories}
We constructed a dataset of academic papers and their corresponding PubMed and arXiv categories. For fair comparison, we collected papers solely from journals that were not seen by \textsc{MIReAD} during training. For PubMed data, we used scientific topic identifiers that come with the journal metadata. For arXiv data, we omitted subcategories and used major categories (e.g. CS.ML and CS.CG were labeled as CS). To ensure that each paper is mapped to a single label, we used arXiv papers with all annotations coming from the same major category. This dataset contains 12K papers across 54 scientific field categories (e.g. physics, computer science, bioinformatics, molecular biology etc.). 

\textbf{Unseen journal classification} This task evaluates whether the learned representations contain very detailed information that allows to distinguish which journal the paper comes from. Since this task resembles \textsc{MIReAD} training objective, we only used journal classes that were not seen during training. This dataset contains the same 12K papers from PubMed and arXiv as the previous task, and 200 journal labels.

We report test set performance of the linear classifier selected by maximal validation set accuracy, and use 4-fold cross validation.

\subsection{Clustering purity}

In our subsequent experiments, we evaluate feature performance when they are used "off-the-shelf", without any finetuning. Such scenario is important for measuring quality of the representations, since it more closely resembles paper search with RS.
\\
Following pre-trained representations assessment strategy from \citet{aharoni2020unsupervised}, we first evaluate clustering using \textit{purity} metric, a widely adopted metric of clustering quality based on intra-cluster similarity \cite{manning2010introduction}. Higher clustering purity indicates model's ability to provide representations that can be more easily grouped into meaningful clusters, such as academic topics. We show results on MAG and MeSH datasets, and perform clustering with k-means algorithm with an increasing number of clusters (10, 20, 50, 100). We compute purity score between ground truth annotations and k-means clustering labels.

\subsection{Information retrieval}
In this final part of our evaluation framework, we measure the quality of representations according to the \textit{information retrieval} perspective. Information retrieval is the process of searching and returning relevant items (in our case scientific documents) based on the input query \cite{manning2010introduction}. For RS, relevant research papers are found based on similarity score between frozen representations. Hence, evaluating how relevant the recommended documents are based on the query document can indicate the quality of the pretrained representations.

For this experiment, we use arXiv subcategories as more stringent labels to measure relevance of representation retrieval \cite{clement2019arxiv}. We collect arXiv papers with their subcategories metadata from six different fields: Computer Science (CS), Mathematics (Math), Physics (Phys), Electrical Engineering and Systems Science (EESS), Economics (Econ) and Statistics (Stat). We perform independent evaluation of subcategories within each field.

We use a commonly adopted evaluation scheme, when pairs of representations are ranked from highest to lowest based on their Pearson's correlation score. Each pair receives a ground truth label of 0 if no subcategories overlap, or 1 otherwise. We report average precision (AP) and area under curve (AUC) scores as final information retrieval metrics.

\begin{table}
\centering
\setlength{\tabcolsep}{0.55em}
\fontsize{9.6}{11}\selectfont
\begin{tabular}{lllll}
\toprule
    & \multicolumn{4}{c}{Number of clusters} \\
    Method $\downarrow$  &    10 &    20 &    50 &   100 \\
\midrule
      BERT & 29.51 & 31.51 &  34.50 & 37.08 \\
PubMedBERT & 32.45 &  32.70 &  37.30 &  40.30 \\
   BioBERT & 33.45 & 35.45 & 41.36 &  45.30 \\
   SciBERT & 29.02 & 31.45 & 35.22 & 38.13 \\
  CiteBERT & 29.22 & 30.53 &  33.90 & 36.73 \\
   SPECTER & 57.28 & \textbf{65.07} & 70.87 & 74.21 \\
    \textsc{MIReAD} & \textbf{57.38} & 64.78 & \textbf{72.15} & \textbf{76.26} \\
\bottomrule
\end{tabular}
\caption{Clustering purity on MeSH dataset with k-means clustering of frozen representations. Results with 10, 20, 50 and 100 clusters across seven methods are reported.}
\label{tab:table2}
\end{table}

\section{Results}
We compared \textsc{MIReAD} with the original BERT model and 5 other approaches that use BERT architecture: PubMedBERT, BioBERT, SciBERT, CiteBERT and SPECTER.

Table  ~\ref{tab:1} shows results of the linear evaluation of representations obtained from seven different models on four tasks/datasets (See Methods).
Overall, \textbf{\textsc{MIReAD} shows a substantial increase in accuracy and F1 score on all four tasks}. On MAG and MeSH tasks \textsc{MIReAD} achieved 84.85\% and 88.22\% accuracy respectively, (81.85 and 86.71 in F1 score). Similarly, \textsc{MIReAD} showed substantial improvement compared to other BERT-based models on 54 PubMed/ArXiv Categories classification and 200 Unseen Journals classification tasks.
\textsc{MIReAD} performance is the closest to SPECTER, although \textsc{MIReAD} outperforms SPECTER in F1 scores across all 4 presented datasets, with statistically significant improvement in 3 cases out of 4. To measure significance of improvement, we performed unpaired t-test between scores of both approaches. The p-values of t-test between F1 scores across 5 runs of SPECTER and \textsc{MIReAD} are 0.2, 0.04, 0.0001 and 0.05, for MAG, MeSH, arxiv \& PubMed, and unseen journals datasets, demonstrating the significant differences for MeSH, arxiv \& PubMed, and unseen journals.

We evaluated the quality of representations with the purity metric of k-means clusters. To compute clustering purity, each cluster is assigned to its “true” class (most frequent class in the cluster), then accuracy is measured by counting the number of correctly assigned documents and dividing by the number of samples. Clustering purity on MeSH (shown in Table~\ref{tab:table2}) and MAG (shown in Appendix~\ref{label:purity_mag}, Table~\ref{tab:sup_table1}) datasets has shown that \textsc{MIReAD} achieves the performance better (on MeSH) or equal (on MAG) to the performance of SPECTER. Both \textsc{MIReAD} and SPECTER significantly outperform all other tested models.

Similar results were obtained on information retrieval experiments with arXiv subcategories (Average Precision is shown in Table  \ref{tab:table3}). Although, SPECTER showed better precision for Math and Physics categories, \textsc{MIReAD} outperformed in Economics,  Computer Sciences (CS) and Electrical Engineering and Systems Science (EESS) categories of arxiv dataset with the improvement of Average Precision of +12.1\% , +11.6\% and +4.7\%, correspondingly.

Overall, three types of evaluations on various datasets reveal that \textbf{\textsc{MIReAD} produces powerful representations of academic papers} whose information content outperforms or matches the performance of the current state-of-the-art feature extraction models.

\begin{table}
\centering
\setlength{\tabcolsep}{0.4em}
\fontsize{8.6}{9}\selectfont
\begin{tabular}{lllllll}
\toprule
    Method &    CS &  Math & Phys &  EESS &  Econ &  Stat \\
\midrule
      BERT & 20.86 & 13.28 & 21.70 & 65.41 & 61.49 &  61.10 \\
PMBERT &  21.00 & 12.54 &   22.81 & 65.79 & 72.05 & 63.36 \\
   BioBERT & 22.98 & 13.07 &   23.26 & 66.28 &  67.40 &  \textbf{64.70} \\
   SciBERT & 23.26 & 14.97 &   21.84 & 67.48 & 64.71 & 62.91 \\
  CiteBERT & 18.75 & 12.59 &    17.50 &  65.70 & 55.74 & 60.47 \\
   SPECTER & 31.97 & \textbf{27.78} &  \textbf{37.17} & 72.53 & 69.66 & 63.91 \\
    \textsc{MIReAD} & \textbf{35.69} & 19.15 &   34.69 & \textbf{75.91} & \textbf{78.12} & 63.99 \\
\bottomrule
\end{tabular}
\caption{Average precision of the representation pairs ranked by correlation scores across arXiv categories.}
\label{tab:table3}
\end{table}

\section{Conclusions}
We present \textsc{MIReAD}, a transformer-based method for representation learning of research articles using minimal information. We fine-tuned \textsc{MIReAD} by predicting the target journal using the paper's title and abstract, and assembled a training dataset spanning over half a million data points from over two thousands journals.
We show that this simple training objective results in high-quality document-level representations, which can be used for various applications and are suitable for recommendation systems.

Earlier we have seen this effect on biological data – where the prediction of subcellular localization (dozens of classes) \cite{razdaibiedina2022learning} or protein (thousands of classes) \cite{razdaibiedina2023pifia} from the fluorescent microscopy images allows to obtain high-quality features. These resulting features had higher information content and could be applied for solving various downstream analysis tasks. Similarly to our findings, more classification labels improved feature quality, which was reflected in downstream task performance.
We found that journal title is a high-quality label for scientific manuscripts, which can be explained by several reasons. Firstly, scientific journals are often highly specialized and focused on a single topic. Therefore, the journal name can serve as a precise topic label. Additionally, journals with different Impact Factors may accept slightly different types of research works, making journal name a valuable human-annotated label.
In our study, the number of journals was determined by available datasets. In a preliminary experiment, we found that increasing the number of labels resulted in better specificity of the representations (data not shown). For example, an increase from 100 to 1000 labels helps the model to learn better separations between sub-fields (e.g.medical sub-domains). We found that lower-level labels encourage the model to learn more fine-grained features to distinguish between journal classes, while high-level labels encourage model to focus on few important features, which may lead to over-simplification of representations content.

Our experimental results show that \textsc{MIReAD} substantially outperforms 6 previous approaches (BERT, PubMedBERT, BioBERT, SciBERT, CiteBERT, SentenceBERT) across three evaluation benchmarks, and outperforms SPECTER, the current SOTA approach for representation learning on scientific articles, in most cases.
The major advantage of \textsc{MIReAD} compared to SPECTER is that \textsc{MIReAD} uses solely paper's abstract and metadata, but does not require the additional information, such as the reference graph. Hence, \textsc{MIReAD} can be trained on novel papers that have not obtained citations or papers that have no open access.

\section{Limitations}
The underlying assumption of our method is that abstract reflects the entire article, creating an unbiased summary of the paper. However, abstract does not guarantee an objective representation of the paper, can often emphasize the main findings while discarding details that the authors deem insignificant. This can lead to potential inaccuracies in paper representations, affecting the results of paper retrieval and recommendation.

Also, in this work we did not exhaust all possible training settings and evaluation strategies due to limited resources. We perform evaluation using three different standards. While we selected the most relevant evaluation tasks, it would be interesting to assess the quality of representations in other ways, such as citation graph reconstruction, predicting reader activity and other clustering-based evaluations. 
Additionally, with the emergence of large-scale language models, another interesting direction for future research is to investigate the relationship between model size and final performance.

\section*{Acknowledgement}
We would like to thank Vector Institute for providing computational resources to run the experiments. Anastasia Razdaibiedina is supported by Vector Institute Postgraduate Affiliate Fellowship.

\bibliography{anthology,custom}
\bibliographystyle{acl_natbib}

\clearpage
\appendix
\section{Appendix}
\label{sec:appendix}

\subsection{Dataset preparation}
\label{label:data_preparation}
\textbf{PubMed}. Since available PubMed datasets did not contain all the necessary metadata, we created a custom dataset by parsing PubMed artciles. We searched PubMed e-utils interface with the custom Python script. The query contained the journal's ISSN and a year of publication. We run through the list of journals from \href{https://www.scimagojr.com}{https://www.scimagojr.com} website and performed searches for years from  2016 to 2021. The list of retrieved PMID then was split into batches of no more than 200 items each and used to download the articles in xml format. The xml page then was parsed for PMID, title, abstract, name of the journal and date of the publication. We only saved articles whose abstracts were written in English to a file. Next, the final list of journals was filtered, such that remaining journals had at least 300 publication in the period of 2016-2021 and at least 1 publication in 2021. For the final dataset, we limited number of articles per journal to 300.
\\
\\
\textbf{arXiv}. We used a dataset of arXiv articles \href{https://huggingface.co/datasets/arxiv\_dataset}{https://huggingface.co/datasets/arxiv\_dataset} available at HuggingFace \cite{wolf2019huggingface}. We limited the number of abstracts per each journal to not exceed 300, and excluded journals with less than 100 abstracts or no publications in 2021. Overall, the arXiv dataset contained >171K abstracts after preprocessing.

\subsection{Computing resources}
We used resources provided by Vector Institute cluster with 528 GPUs,
6 GPU nodes of 8 x Titan X, and 60 GPU nodes each with 8 x T4, for development and deployment of large-scale transformer-based NLP models.

\subsection{Linear probing experiments}
\label{label:linear_probing}
For our linear probing experiments, we used multinomial logistic regression with a learning rate of 5e-4 and batch size of 100, which we trained for 5 epochs. We did not add a regularization penalty as we found that the regression model did not overfit due to its simplicity. We used 4-fold cross-validation with early stopping based on the maximal validation set performance, and our final performance is averaged across all cross-validation runs.

\subsection{Clustering purity on MAG dataset}
\label{label:purity_mag}
We include results for clustering purity experiments on MAG datset in Table~\ref{tab:sup_table1}.
\begin{table}[h]
\setlength{\tabcolsep}{0.55em}
\fontsize{9.6}{11}\selectfont
\begin{tabular}{lllll}
\toprule
    Method &    10 &    20 &    50 &   100 \\
\midrule
      BERT & 38.92 & 50.94 & 57.49 & 60.43 \\
PubMedBERT & 31.64 & 47.49 & 58.41 & 60.48 \\
   BioBERT & 43.44 & 56.38 & 61.63 & 65.27 \\
   SciBERT & 46.98 & 48.83 & 57.35 & 60.11 \\
  CiteBERT &  33.9 & 45.05 & 51.73 & 55.55 \\
   SPECTER & 61.95 & 75.03 & 78.07 & 78.67 \\
    MIReAD & 61.03 &  71.9 & 75.31 & 78.63 \\
\bottomrule
\end{tabular}
\caption{Clustering purity on MAG dataset with k-means clustering of frozen representations. Results with 10, 20, 50 and 100 clusters across seven methods are reported.}
\label{tab:sup_table1}
\end{table}

\subsection{arXiv subcategories}
Table~\ref{tab:sup_table2} includes a description of arXiv subcategories that we used to form a category for article topic classification. 

\begin{table*}
\centering
\setlength{\tabcolsep}{0.4em}
\fontsize{8.6}{9}\selectfont
\begin{tabular}{p{0.18\linewidth} | p{0.025\linewidth}| p{0.65\linewidth}}
\toprule
    Categories & \#\# & Subcategories \\
\midrule
      Computer Science (CS) & 40 & Artificial Intelligence, Hardware Architecture, Computational Complexity, Computational Engineering, Finance, and Science, Computational Geometry, Computation and Language, Cryptography and Security, Computer Vision and Pattern Recognition, Computers and Society, Databases, Distributed, Parallel, and Cluster Computing, Digital Libraries, Discrete Mathematics, Data Structures and Algorithms, Emerging Technologies, Formal Languages and Automata Theory, General Literature, Graphics, Computer Science and Game Theory, Human-Computer Interaction, Information Retrieval, Information Theory, Machine Learning, Logic in Computer Science, Multiagent Systems, Multimedia, Mathematical Software, Numerical Analysis, Neural and Evolutionary Computing, Networking and Internet Architecture, Other Computer Science, Operating Systems, Performance, Programming Languages, Robotics, Symbolic Computation, Sound, Software Engineering, Social and Information Networks, Systems and Control \\ \midrule

   Mathematics (Math) & 32 & Commutative Algebra, Algebraic Geometry, Analysis of PDEs, Algebraic Topology, Classical Analysis and ODEs, Combinatorics, Category Theory, Complex Variables, Differential Geometry, Dynamical Systems, Functional Analysis, General Mathematics, General Topology, Group Theory, Geometric Topology, History and Overview, Information Theory, K-Theory and Homology, Logic, Metric Geometry, Mathematical Physics, Numerical Analysis, Number Theory, Operator Algebras, Optimization and Control, Probability, Quantum Algebra, Rings and Algebras, Representation Theory, Symplectic Geometry, Spectral Theory, Statistics Theory \\ \midrule
  Physics (Phys) & 51 & Cosmology and Nongalactic Astrophysics, Earth and Planetary Astrophysics, Astrophysics of Galaxies, High Energy Astrophysical Phenomena, Instrumentation and Methods for Astrophysics, Solar and Stellar Astrophysics, Disordered Systems and Neural Networks, Mesoscale and Nanoscale Physics, Materials Science, Other Condensed Matter, Quantum Gases, Soft Condensed Matter, Statistical Mechanics, Strongly Correlated Electrons, Superconductivity, General Relativity and Quantum Cosmology, High Energy Physics - Experiment, High Energy Physics - Lattice, High Energy Physics - Phenomenology, High Energy Physics - Theory, Mathematical Physics, Adaptation and Self-Organizing Systems, Chaotic Dynamics, Cellular Automata and Lattice Gases, Pattern Formation and Solitons, Exactly Solvable and Integrable Systems, Nuclear Experiment, Nuclear Theory, Accelerator Physics, Atmospheric and Oceanic Physics, Applied Physics, Atomic and Molecular Clusters, Atomic Physics, Biological Physics, Chemical Physics, Classical Physics, Computational Physics, Data Analysis, Statistics and Probability, Physics Education, Fluid Dynamics, General Physics, Geophysics, History and Philosophy of Physics, Instrumentation and Detectors, Medical Physics, Optics, Plasma Physics, Popular Physics, Physics and Society, Space Physics, Quantum Physics \\ \midrule
  Electrical Engineering and Systems Science (EESS) & 4 & Audio and Speech Processing, Image and Video Processing, Signal Processing, Systems and Control \\ \midrule
  Economics (Econ) &  3 & Econometrics, General Economics, Theoretical Economics \\ \midrule
  Statistics (Stat) & 6 & Applications, Computation, Methodology, Machine Learning, Other Statistics, Statistics Theory \\
\bottomrule
\end{tabular}
\caption{arXiv categories.}
\label{tab:sup_table2}
\end{table*}

\end{document}